\pdfoutput=1

\documentclass[a4paper,fleqn]{cas-dc}

\usepackage{algorithmic}
\usepackage{array}
\usepackage[caption=false,font=footnotesize,labelfont=rm,textfont=rm]{subfig}
\usepackage{stfloats}
\usepackage{url}
\usepackage[authoryear,longnamesfirst]{natbib}
\usepackage{amsmath,amsfonts}
\usepackage{graphicx}
\usepackage{amsmath}
\usepackage{amssymb}
\usepackage{picinpar}
\usepackage{listings}
\usepackage[boxed]{algorithm2e}
\usepackage{color}
\usepackage{overpic}
\usepackage{verbatim}  
\usepackage{textcomp}
\usepackage{cuted}
\usepackage{float}
\usepackage{pgfplots}
\pgfplotsset{compat=1.18}
\usepackage{dsfont}
\usepackage{multirow}
\usepackage[english]{babel}

\def\tsc#1{\csdef{#1}{\textsc{\lowercase{#1}}\xspace}}
\tsc{WGM}
\tsc{QE}

\begin{document}
\let\WriteBookmarks\relax
\def\floatpagepagefraction{1}
\def\textpagefraction{.001}



  \title[mode = title]{Edge Based Oriented Object Detection}

  \tnotemark[1,2]  

  \tnotetext[1]{This document is the results of the research project
    funded by the National Science Foundation.} 

  \tnotetext[2]{The second title footnote which is a longer text
    matter to fill through the whole text width and overflow into
    another line in the footnotes area of the first page.}

  \author[1]{Jianghu Shen}[type=editor,
    orcid=0009-0005-9302-2165]
  \fnmark[1] 
  \ead{21s053069@stu.hit.edu.cn} 

  \credit{Conceptualization of this study, Methodology, Software} 

  \address[1]{School of Mechanical Engineering and Automation, Harbin Institute of Technology Shenzhen, Shenzhen, 518055, China} 

  \author[1]{Xiaojun Wu}[]
  \cormark[1]  
  \fnmark[2] 
  \ead{wuxj@hit.edu.cn}
    





  \cortext[cor1]{Corresponding author} 
  \fntext[fn1]{This is the first author footnote.}

  \fntext[fn2]{Another author footnote, this is a very long footnote and it should be a really long footnote. But this footnote is not yet sufficiently long enough to make two lines of footnote text.}



\begin{abstract}
In the field of remote sensing, we often utilize oriented bounding boxes (OBB) to bound the objects. This approach significantly reduces the overlap among dense detection boxes and minimizes the inclusion of background content within the bounding boxes. To enhance the detection accuracy of oriented objects, we propose a unique loss function based on edge gradients, inspired by the similarity measurement function used in template matching task. During this process, we address the issues of non-differentiability of the  function and the semantic alignment between gradient vectors in ground truth (GT) boxes and predicted boxes (PB). Experimental results show that our proposed loss function achieves $0.6\%$ mAP improvement compared to the commonly used Smooth L1 loss in the baseline algorithm. Additionally, we design an edge-based self-attention module to encourage the detection network to focus more on the object edges. Leveraging these two innovations, we achieve a mAP increase of $1.3\%$ on the DOTA dataset.
\end{abstract}




\maketitle

\section{Introduction}
\label{sec:intro}
With the development of deep learning, object detection, as a fundamental task in computer vision, has achieved higher detection accuracy. However, in certain special scenarios such as aerial remote sensing, text detection, and e-commerce logistics, where the interested targets can be oriented in arbitrary directions, using conventional horizontal bounding boxes (HBB) to represent objects can introduce a series of issues. These issues include bounding box overlap, excessive inclusion of background information within the bounding box, and the shape of the bounding box changing with the angle of the target \cite{yin2018fast}.

To address these problems, the task of oriented object detection has been proposed. Unlike traditional horizontal object detection tasks, this task employs rotated bounding boxes with an angle parameter $\theta$ to more accurately enclose the targets. As a result, numerous algorithms focused on improving detection accuracy have been proposed. Similar to horizontal object detection algorithms, they can generally be classified into one-stage and two-stage methods. Two-stage methods, such as \cite{xie2021oriented}, \cite{ding2019learning} \cite{han2021redet}, firstly, generating a series of proposal boxes that likely contain the target by region proposal networks (RPN). Then, followed by more refined regression and classification of these proposal boxes. On the other hand, one-stage methods, such as \cite{yang2021r3det} \cite{han2021align} \cite{lin2017focal}, simplify this process by directly generating final prediction box for each targets and their corresponding classes. Typically, one-stage methods offer advantages in terms of training and inference speed but may not achieve the same level of detection accuracy as two-stage methods.

\begin{figure}[ht]
	\centering
    \includegraphics[width = 0.9\linewidth]{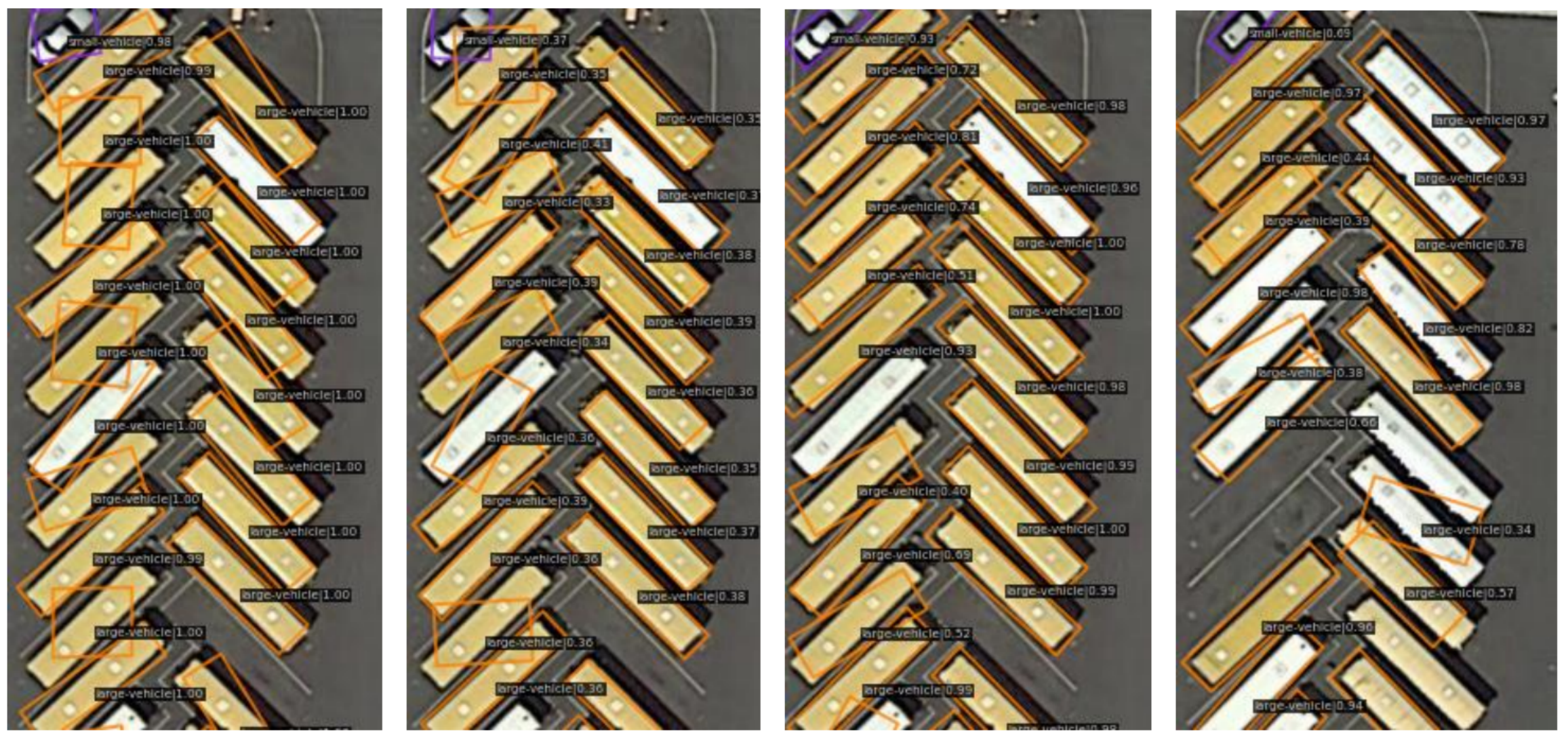}
	\caption{The performance of some deep learning methods for oriented object detection on the DOTA dataset is not satisfactory, especially in terms of angle estimation accuracy.}
	\label{fig:other algorithm}       
\end{figure}

Despite the continuous improvements on accuracy achieved by these oriented object detection algorithms, we have identified a remaining limitation in their ability to accurately detect the angles of targets. Fig. \ref{fig:other algorithm} shows the detection results of several classical algorithms on the DOTA dataset, revealing significant angle detection errors for a considerable number of targets. However, in another subfield of traditional computer vision, template matching, we have noted that shape-based template matching methods exhibit exceptional accuracy in angle detection. After subpixel correction, the angle error can typically be reduced to the level of 0.01°. This remarkable performance is mainly attributed to the edge extraction performed on the template and source images before matching, as well as the utilization of edge gradient vectors rather than pixel grayscale values for similarity calculation. Edge gradient vectors possess strong directionality, which contributes to improved angle detection accuracy. Therefore, we have adapted and modified a certain shape-based template matching method for integrating into deep learning-based oriented object detection task. Additionally, we have developed a novel loss based on edge gradient vectors and a self-attention mechanism focusing on the edges. The objective of these developments is to enhance the detection network's ability to achieve higher detection accuracy for oriented objects, specifically aiming for higher mAP and lower angle errors.

Specifically, our work incorporates the following two innovations:

(1) A special regression loss based on edge gradient vectors was proposed  and it was modified from the similarity function used in shape-based template matching, addressing the issues of non-differentiability when used as a loss function and the semantic misalignment between edge gradient vectors in PB and GT boxes. We name this loss function "Edge-Loss". As is well-known, the fundamental paradigm of template matching involves sliding a window of the same size as the template over the source image. The similarity between the local image within the sliding window and the template is calculated by the similarity function, and the position(s) with the highest similarity are selected as the matching result. In the deep learning-based oriented object detection task, we treat GT as the template and PB as the sliding window, using the extracted edge gradient vectors to compute the similarity between GT and PB. The specific calculation method will be detailed in Section \ref{sec:method}. By replacing the commonly used smooth L1 Loss with Edge-Loss, we have achieved significant improvements in the mAP evaluation metric on DOTA dataset.

(2) We have designed an edge self-attention module based on the edge information of images, aiming to encourage the detection network to focus more on object edges. Currently, many scholars have applied self-attention mechanisms in object detection networks, such as channel self-attention \cite{hu2018squeeze} \cite{wang2020eca}, spatial self-attention \cite{jaderberg2015spatial} \cite{hu2018gather}, and kernel selection mechanism \cite{li2019selective} \cite{li2023large}. However, no previous researchers have incorporated edge information into self-attention mechanisms. In human perception, object edges play a significant role in object recognition, particularly for regular objects like vehicles, playgrounds, and tennis courts. The straight edges often play a decisive role in determining the object's orientation. Therefore, we have developed an Edge self-attention module and applied it to the Feature Pyramid Network (FPN) to enhance the edge information of each level of feature maps. The improved FPN is named Edge-FPN. By replacing the traditional FPN with Edge-FPN, we have also achieved performance improvements.

\section{Related Work}
\label{sec:relwork}
\subsection{Template Matching}
The loss function we designed is closely related to the similarity function used in template matching. 
Currently, based on different measuring criteria, template matching can be categorized into the following three types:

1) Gray-level-based template matching methods, such as SSD \cite{overview2016}, SAD, and NCC \cite{ncc}. These methods measure the similarity by comparing gray-level differences pixel by pixel between the subimage within the sliding window and the template. While these methods have a high matching speed, they have poor robustness. Factors such as illumination variations, noise, and blur can significantly affect the gray-level values of local pixels.

2) Methods based on the nearest neighbor field. These methods perform matching based on the mapped point sets rather than pixel-level matching. Examples of such methods include BBS \cite{dekel2015best}, DDIS \cite{talmi2017template}, DIWU \cite{talker2018efficient} and GAD \cite{lan2021gad}. These methods first map the subimages of the template and the original image to a higher-dimensional space and generate two sets of points, $P$ and $Q$. Then, they search for the closest point pairs $(p_i, q_j)$ between $P$ and $Q$, using the number of point pairs as the similarity indicator. This type of method has the fastest matching speed and does not require strict shape and texture similarity between the template and the objects. However, these methods often only provide an approximate location of the objects of interest and have lower matching accuracy.

3)Shape-based methods, which utilize the geometric features of objects for analysis and recognition \cite{steger2002occlusion} \cite{lu2021robust}. The loss function we designed is developed based on the edge-based similarity of this category, the specific formula for the similarity function is as follows: 
\begin{equation}\label{eqn:1}
s\left(r,c\right)=\frac{1}{n}\sum_{i=1}^{n}\frac{\langle \boldsymbol{p}_i^m, \boldsymbol{p}_{\left(r+r_i,c+c_i\right)}^s  \rangle}{\| \boldsymbol{p}_i^m\| \cdot \| \boldsymbol{p}_{\left(r+r_i,c+c_i\right)}^s\|}
\end{equation}
In the equation, $p_i^m$ represents the edge gradient vector in the template, $p_i^s$ represents the gradient vector at the same position in the sliding window, ||·|| denotes the Euclidean norm, and <·> is the dot product of two vectors. This type of method exhibits high accuracy and is less sensitive to disturbances such as illumination changes, noise, blur, and occlusion. However, it has a longer matching time. Recently, many scholars devoted to improving the matching efficiency of such algorithms and reducing computational time. For example, Shen (citing after published) proposed a shape-based template matching acceleration algorithm that achieves matching times comparable to NCC and other gray-based methods.

\subsection{Oriented object detection}

Like horizontal object detection algorithms, oriented object detection algorithms can also be roughly divided into two categories: one-stage algorithms and two-stage algorithms, with most of these algorithms having prototypes in horizontal object detection. Two-stage algorithms, such as RoI transformer \cite{ding2019learning}, oriented RCNN \cite{xie2021oriented}, SCRDet \cite{yang2019scrdet}, KFIoU \cite{yang2021learning}, etc., generate proposal boxes in the first stage, also known as the region proposal network (RPN). In this stage, preliminary binary classification is performed on predefined dense anchor boxes, dividing them into foreground and background boxes, as well as regression. Then, some foreground and background boxes are fed into the second stage for the second round of regression and more specific classification. Generally, between these two stages, an alignment operation is also performed to align local features within proposal boxes of different sizes to a fixed-dimensional fully connected layer, such as RoI Align and RoI pooling. Although two-stage algorithms achieve higher performance on benchmarks dataset, such as DOTA \cite{xia2018dota}, CoCo \cite{lin2014microsoft}, and HRSC2016 \cite{liu2016ship}, they are often too slow to be applied to real-time detecting tasks. On the contrary, one-stage oriented object detection frameworks, for example, S2ANet \cite{han2021align}, R3Det \cite{yang2021r3det}, DRN \cite{pan2020dynamic}, obtain the final predicted bounding boxes through a single classification and regression step from anchor boxes, resulting in faster detection speed but generally lower accuracy compared to two-stage algorithms.

Although oriented object detection shares the same pipeline as horizontal object detection, with an additional regression parameter, the angle $\theta$ of the bounding box, there are also some sepcial issues to be noted in oriented object detection tasks. The angle periodicity (PoA) and square-like detection problem (SLP) cause the predicted boxes to approach the corresponding GT boxes in a complex and convoluted path during the training process. Currently, many methods have been proposed to address these problems. For example, \cite{yang2021rethinking} models the oriented bounding boxes as Gaussian distributions and approximates the IoU loss using the Gaussian Wasserstein distance (GWD). Because GWD is insensitive to the definition of bounding boxes, it elegantly solves the aforementioned problems. Currently, there are also anchor-free algorithms, such as Oriented RepPoints \cite{li2022oriented}, which locates the interested objects by regressing nine key points of the target. Additionally, due to the presence of the angle parameter in the bounding box, computing the SkewIoU of two oriented rectangular boxes in differentiable ways becomes challenging. PIoU \cite{chen2020piou} approximates each pixel's position and calculates SkewIoU by counting the number of pixels inside the box. Another method \cite{zheng2020rotation} calculates $IoU_1$ by fixing bounding box $A$ and rotating bounding box $B$ to be perpendicular to A, then calculates the horizontal bounding box $IoU$ between $A and $B, and exchanges the roles of $A and $B to obtain $IoU_2$. Finally, $min{IoU_1, IoU_2}$ is used as the final SkewIoU. KFIoU \\cite{yang2021learning} uses the Gaussian distribution modeling of the overlapped area of rectangular boxes after being filtered with Kalman filtering to simulate the definition of SkewIoU and proves that the trend of KFIoU is consistent with SkewIoU.

In addition, much research has focused on the design of self-attention modules in detection networks. SENet \cite{hu2018squeeze} use the global average pooling value of feature maps to reweight the channels, realizing channel self-attention mechanism. Spatial self-attention mechanisms, such as GCNet \cite{lin2021global} and GENet \cite{hu2018gather}, enhance certain local regions of feature maps through spatial masks. SKNet \cite{li2019selective} and LSKNet \cite{li2023large} perform adaptive selection between the convolution results of different-sized kernels.

In this paper, we first approximate the position of the edge gradient vector using a method similar to PIoU and design a differentiable loss based on edge information. Then, we utilize the edge mask feature map for spatial self-attention mechanism, simulating human's behavior of differentiating foreground and background based on the edges of objects in the real world. Detailed algorithms and network designs will be presented in Section \ref{sec:method}.

\section{Methodology}
\label{sec:method}
\begin{figure*}[ht]
	\centering
    \includegraphics[width = 0.8\linewidth]{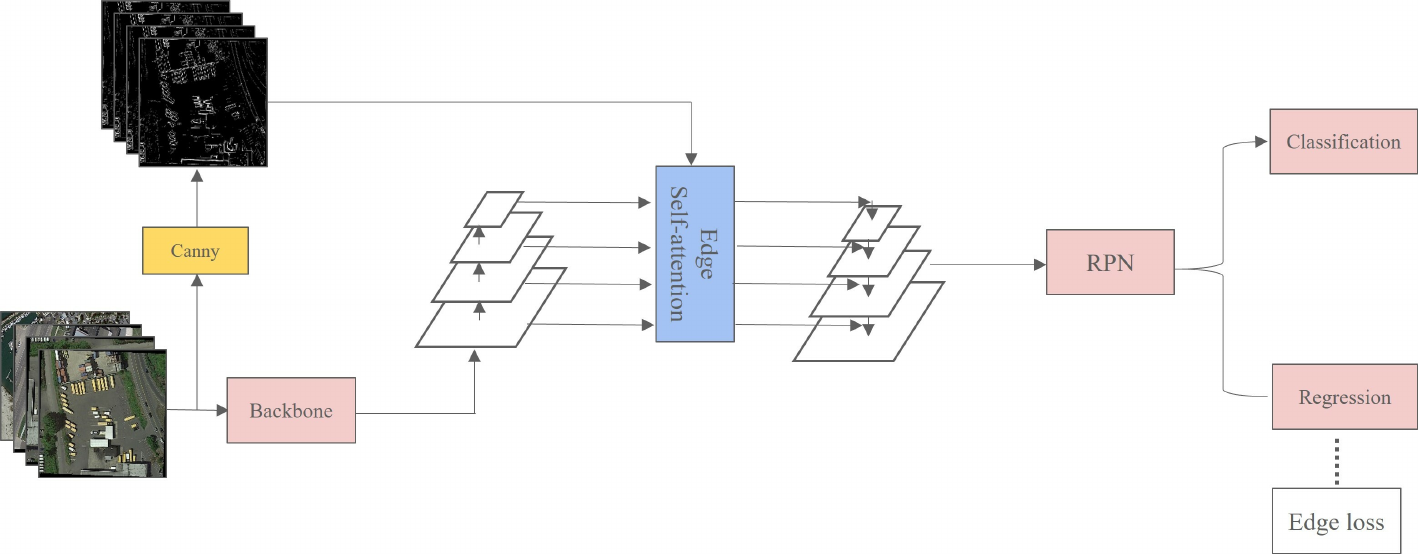}
	\caption{Construction of our network}
	\label{fig:construction}       
\end{figure*}

The architecture of our algorithm is shown in Fig. \ref{fig:construction}. Firstly, we enhance the top-down process in the FPN by incorporating edge information from the source image. The modified FPN is referred to as Edge-FPN, where the pixels corresponding to the edges in the feature maps are enhanced, while the non-edge regions are suppressed. In Section \ref{sec:experiment}, we also experiment with a fusion approach that combines Edge-FPN with the traditional FPN, resulting in improved performance. The Edge-Loss, based on edge gradient vectors, is employed in the regression branch of the detection head, replacing the commonly used Smooth L1 Loss.

\subsection{Edge Self-attention mechanism}
\begin{figure*}[ht]
	\centering
    \includegraphics[width = 0.8\linewidth]{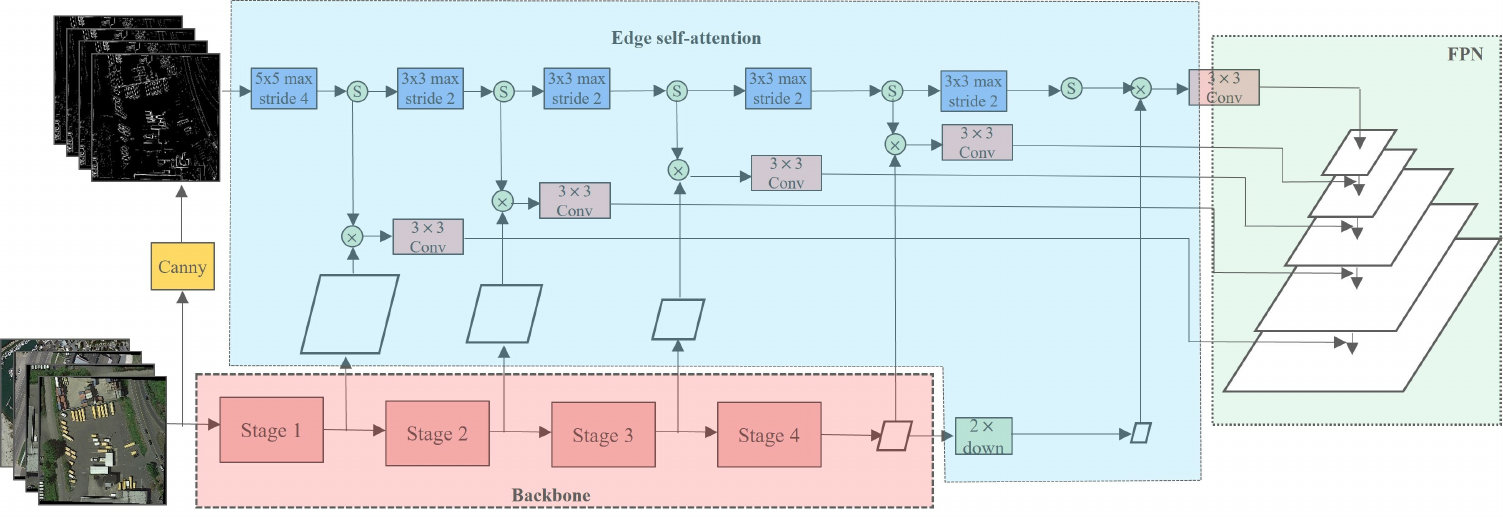}
	\caption{Construction of the edge self-attention}
	\label{fig:self-attention}       
\end{figure*}

To enhance the detection network's ability to focus on object edges, we designed the Edge-FPN architecture shown in Fig. \ref{fig:self-attention}. Firstly, we collect the results from each stage of the backbone. Taking Oriented RCNN as an example, we downsample the output of the last stage of ResNet $F_4$ by a factor of 0.5 to generate $F_5$.

\begin{equation}\label{eqn:52}
F_i=\left[F_1 ; \ldots ; F_5\right]
\end{equation}

where the spatial resolution of $F_i$ is half that of $F_{i-1}$.


Then, We use the Canny operator to extract the edges of the source image. To ensure the sufficient number of edge vectors, we employ adaptive high and low thresholds for the Canny operator. The initial values of the high threshold and low threshold are set to 80 and 40, respectively. After each time of edge extraction, we count the number of edge gradients. If the count is less than 0.05 times of the area of the source image, the high and low thresholds will be updated by multiplying 0.7. We repeat this process of edge extraction using the adjusted high and low thresholds until the desired number of gradient vectors is obtained. This resulting edge map is denoted as $E_0$.

Next, we apply the sigmoid function to compress the lengths of the edge gradients into the range of $[0, 1]$. Then, we perform multiple rounds of max pooling on $E_0$ to obtain:

\begin{equation}\label{eqn:2}
E_i=\left[E_1 ; \ldots ; E_5\right]
\end{equation}
Where each $E_i$ and $F_i$ have the same spatial resolution.

The final output of the edge self-attention mechanism is the element-wise product between feature map F and edge map E, as:
\begin{equation}\label{eqn:4}
Y = E \cdot F
\end{equation}
Finally, the feature maps Y are used for the top-down construction of the edge-FPN.

\subsection{Edge Loss}
A phenomenon have been observed in shape-based template matching tasks, if the templates and the interested targets have the same shape, the extremely high accuracy of angle detection could be obtained. It mainly because of directional edge vectors utilized in shape-based template matching. However, it fails to provide reliable matching results when the target's shape vary significantly. Therefore, in this study, we attempt to incorporate the similarity function used in shape-based template matching into deep learning-based oriented object detection to address the issue of insufficient angle detection accuracy in deep learning-based methods. The formula for the shape-based similarity function is shown in Eq. \ref{eqn:1}, which measures the similarity between the template and the local image within the sliding window on the source image. similarly, we use the sub-images inside the GT boxes as templates and treat the PB as sliding window in the deep learning-based oriented object detection task, and calculate the similarity between the local images within the GT and PB.

\subsubsection{Position approximation of the edge gradient vectors}
However, since the formula \ref{eqn:1} does not include the five parameters $(x, y, w, h, \theta)$ of PB, it is not possible to regress the proposal boxes during the training process.
To address this issue, we refer to the method used in PIoU \cite{chen2020piou} to approximate whether each edge gradient vector is inside the specified bounding box.
The equation for approximating is as follow:

\begin{equation}\label{eqn:5}
    \begin{aligned}
& &a\left(g_i \mid b\right)=G\left(d_i^w, b_w\right) \cdot G\left(d_i^h, b_h\right) \\
& where \\
& &G\left(d_i^w, b_w\right)=1-\frac{1}{1+e^{-10\left(d_i^w-b_w\right)}}
\end{aligned}
\end{equation}

where $g_i$ represents the edge gradient vector, $b$ represents a bounding box, $b_w$ and $b_h$ are the width and height of the bounding box, and $d_i^w$ denotes the distance between the $g_i$ to the short axis of $b$, while $d_i^h$ represents the distance of $g_i$ to the long axis.

\begin{figure}[ht]
	\centering
    \includegraphics[width = 0.8\linewidth]{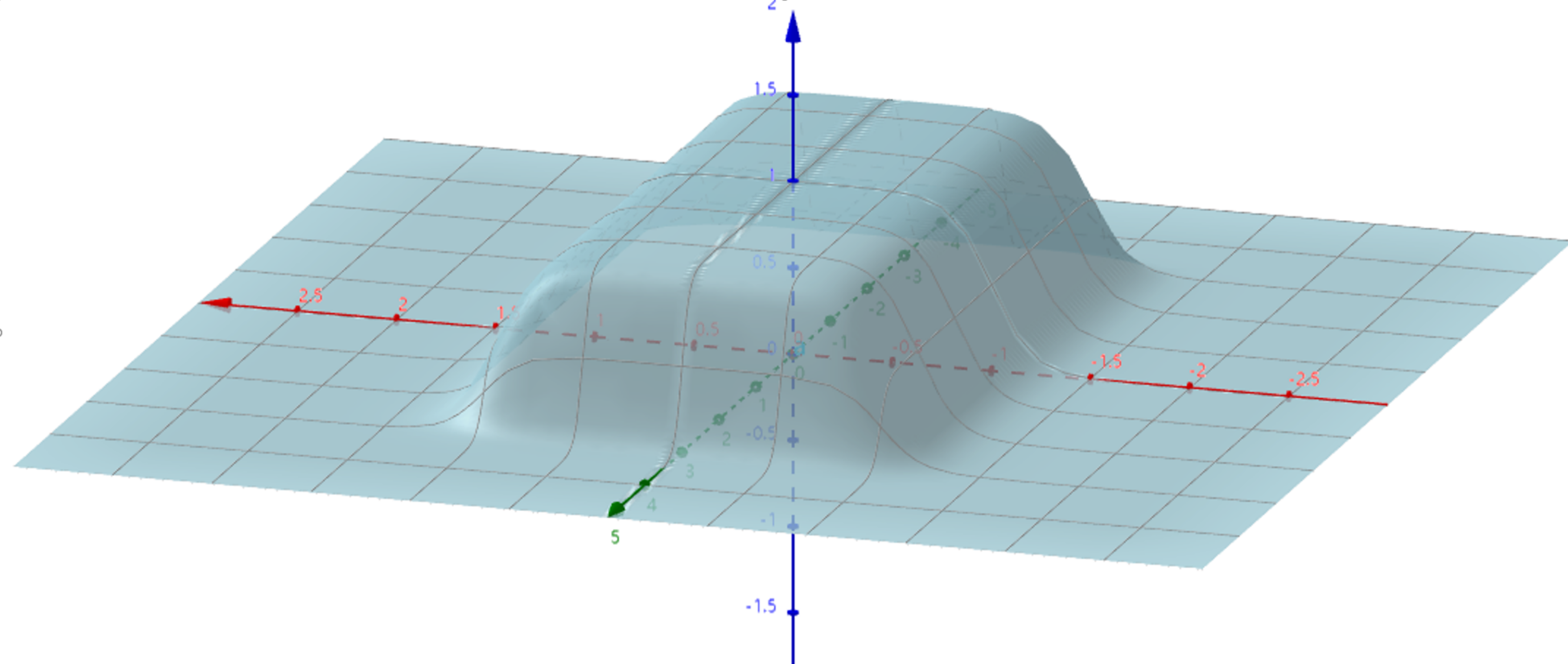}
	\caption{The visualization of the position approximation function}
	\label{fig:approximately function}       
\end{figure}

The visualization of this function is shown in Fig. \ref{fig:approximately function}, and it has the following three characteristics:

1) If the gradient $g_i$ is inside the bounding box, the function value is close to 1.

2) If the gradient $g_i$ is outside the bounding box, the function value is close to 0.

3) At the boundary of the bounding box, the function value smoothly decreases from 1 to 0.

By weighting each gradient vector using the function \ref{eqn:5}, we approximate the retention of vectors inside the bounding box, and rejecting the outside vectors. Essentially, the calculation of this function involves the regression parameters $(x, y, w, h, \theta)$ of the PB. This allows the formula \ref{eqn:1} to have the potential to become a loss function.

\subsubsection{Semantic Error of Edge Gradient Vectors between GT and PB}
In template matching tasks, we can ensure that the template and the sliding window have the same size, which guarantees that each pair of gradient vectors in the sliding window and template have the same semantics when the window slides onto the actual target. However, in deep learning-based object detection tasks, the PB is constantly undergoing dynamic adjustments, and it cannot be guaranteed to have the same shape and position as the GT. Therefore, we use relative positions to search for gradient vectors within PB that match the gradient vectors inside GT box.

\begin{figure}[ht]
    \centering
    \includegraphics[width = 0.8\linewidth]{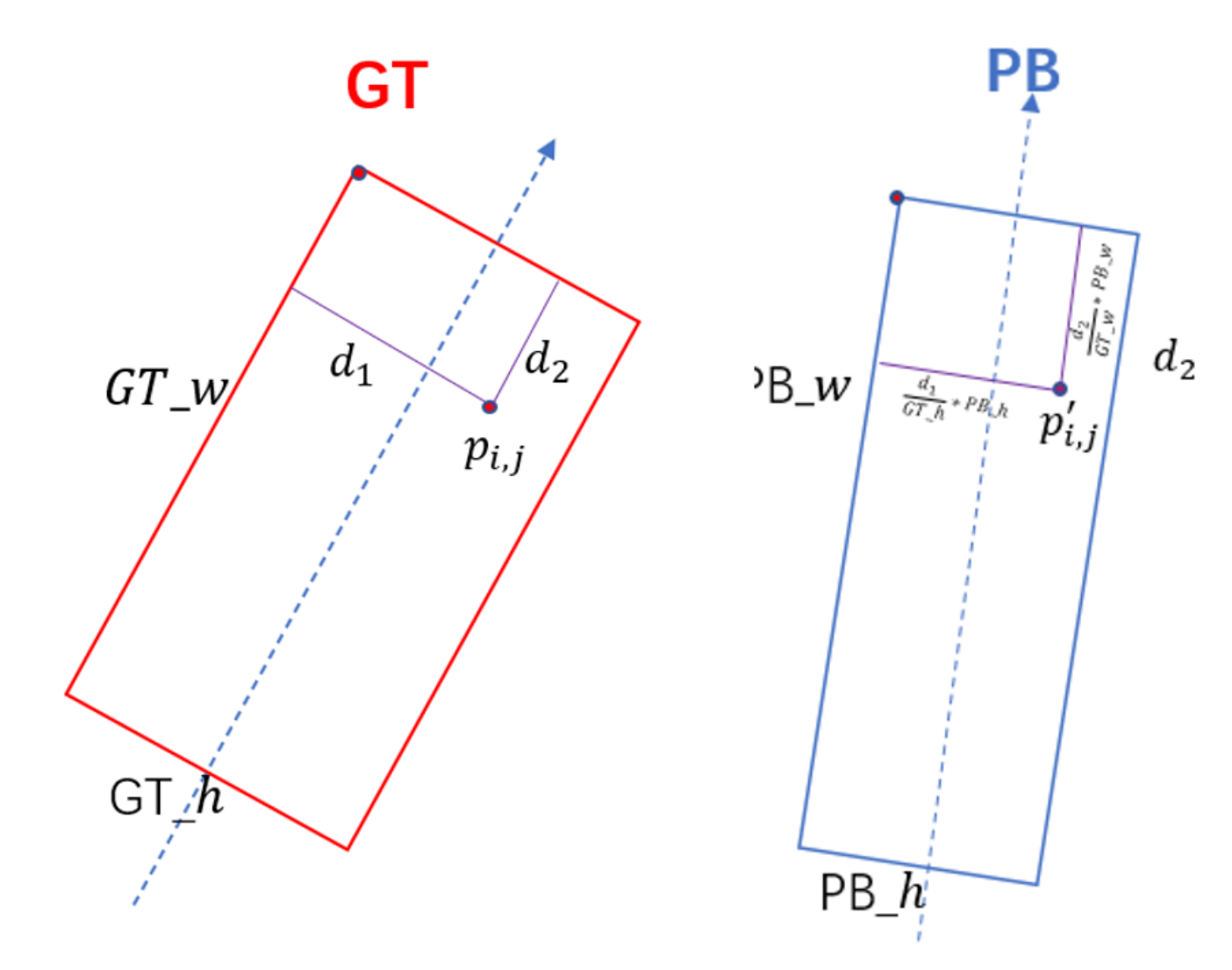}
    \caption{The position relationship of gradient vectors in GT box and PB}
    \label{fig:relative_position}
\end{figure}

First, we take the top-left corner of the GT box as the reference position and determine the distances of each gradient vector within the bounding box to the two edges connected to the top-left corner, denoted as $(d_1, d_2)$. Then we could get the relative distances $(d_1/GT_h, d_2/GT_w)$ with respect to the GT box. These relative distances are applied to the corresponding PB, resulting in the position in PB that correspond to the edge gradient vector in the GT box, denoted as $(PB_hd_1/GT_h, PB_wd_2/GT_w)$. These coordinates are also refer to the top-left corner of PB. The position relationship of these gradient vectors is illustrated in Fig. \ref{fig:relative_position}.

\begin{figure}[ht]
    \centering
    \includegraphics[width = 0.8\linewidth]{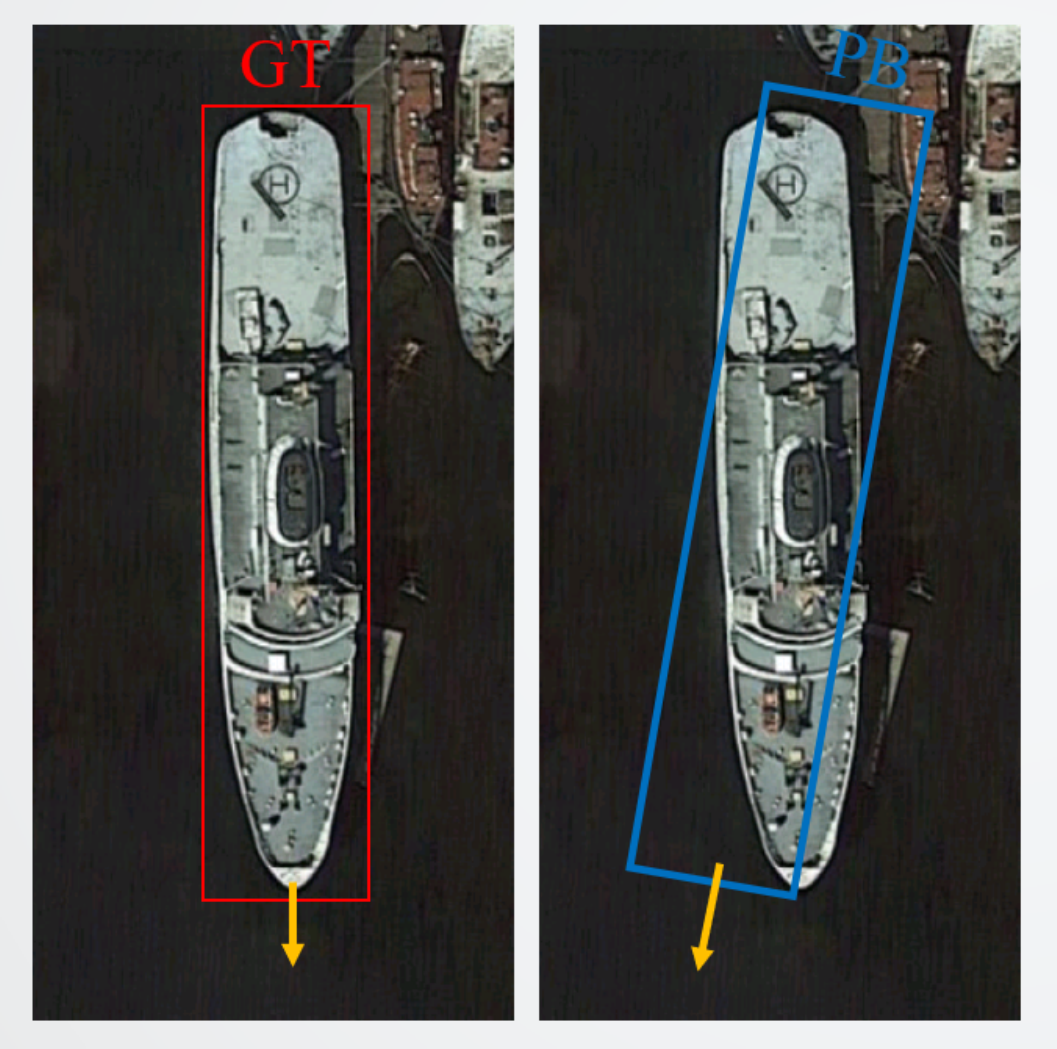}
    \caption{The corresponding position in PB can introduce semantic errors, where the vector don't precisely align with the bow of the ship.}
    \label{fig:semantic error}
\end{figure}

But a issue still exists, that is the matched gradient vectors obtained using the aforementioned approach may appear at different semantic positions from the true object. For example, as shown in Fig. \ref{fig:semantic error}, if we find a gradient vector at the bow of the ship within the GT box, the corresponding position in PB obtained by above method may deviate from the actual position of the bow. We refer to this error as semantic error. To mitigate the performance degradation caused by this error, we propose two prior knowledge principles:

1) We assume that the semantic error between the gradient vectors at corresponding positions in the GT box and PB is small, witch means the obtained gradient vector in PB should be near where it was supposed to be.

2) We assume that the semantic error follows the Gaussian distribution.

To mitigate the semantic errors caused by the mismatch between the PB and GT positions, the proposed Algorithm \ref{alg:1} employs a strategy that considers multiple gradient vectors in the vicinity of the corresponding PB position, instead of relying on a single vector. Additionally, a two-dimensional Gaussian distribution is utilized to weight these gradient vectors.

\begin{algorithm}
	\renewcommand{\algorithmicrequire}{\textbf{Require:}}
	\caption{Get Gradient weight for the vectors in the vicinity of the corresponding PB position}
	\label{alg:1}
	\begin{algorithmic}[1]
        \STATE Initialization: $GT, PB, edge\_images, weigth \leftarrow [] $
		\FOR{point $p_i$ in $edge\_images$}
        \STATE Get distances $d_1, d_2$ from two sides contacted with top-left corner of the GT Box
        \STATE Get relative distances $d_1/GT_h, d_2/GT_w$
        \STATE Find position $p^{'}_i$ at $(d_1*PB_h/GT_h, d_2*PB_w/GT_w)$
            \FOR{point $p_j$ in $edge\_images$}
                \STATE get distance $d_{i,j}$ between ($p^{'}_i,p_j$)
                \STATE $get weight Guassian({d_i,j})$
            \ENDFOR
        \ENDFOR
        \ENSURE $weight$
	\end{algorithmic}  
\end{algorithm}

In conclusion, the final formulation of the edge-Loss, as shown in Eq. \ref{eqn:6}, addresses both the issue of non-differentiability of the regression parameters and mitigates the impact of semantic errors caused by differences of the positions and shapes between PB and GT.

\begin{equation}\label{eqn:6}
\operatorname{Loss}=\frac{1}{n^2} \sum_{i=1}^n \sum_{j=1}^n w_{i, j} g_i g_j^{\prime} a\left(g_i \mid GT\right) a\left(g_j^{\prime} \mid PB\right)
\end{equation}

where the $g_i$ represents the gradient vector in the GT box, and the $g_j^{\prime}$ represents the gradient vector at the corresponding position in the PB. The $w_{i,j}$ signifies the Gaussian weight of the gradient vector $g_j^{\prime}$. The terms $a\left(g_i \mid GT\right)$ and $aa\left(g_j^{\prime} \mid PB\right)$ indicate whether the gradient vectors $g_i$ and $g_j^{\prime}$ are in their respective GT box and PB.

\section{Experiments Results}
\label{sec:experiment}

\subsection{Dataset and implementation details}
\label{subsec:dataset}

\begin{figure}[ht]
    \centering
    \includegraphics[width = 0.8\linewidth]{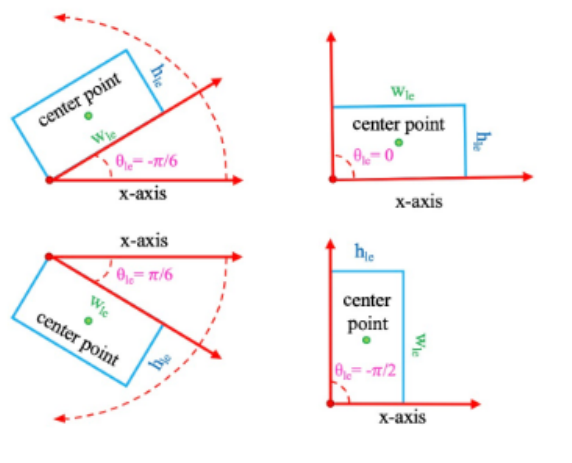}
    \caption{The definition of LE90 annotation format}
    \label{fig:LE90}
\end{figure}

We evaluated the performance of our algorithm on the DOTA-v1.0 dataset \cite{xia2018dota}. It consists of 2,806 aerial images with varying sizes ranging from 800x800 to 4,000x4,000 pixels, and a total of 188,282 fully annotated instances across 15 different classes, including plane (PL), ship (SH), storage tank (ST), baseball diamond (BD), tennis court (TC), basketball court (BC), ground track field (GTF), harbor (HA), bridge (BR), large vehicle (LV), small vehicle (SV), helicopter (HC), roundabout (RA), soccer ball field (SBF) and swimming pool (SP).In the dataset, ST and RA are annotated using horizontal rectangular boxes, while the other object categories are annotated via oriented rectangular boxes. The bounding box in the dataset are represented by the four corner points $(x1, y1, x2, y2, x3, y3, x4, y4)$ arranged in clockwise order. When using the DOTA-v1.0 dataset, we transformed the bounding box annotations to the Long Edge Definition 90° (LE90) annotation format, The definition is illustrated in Fig \ref{fig:LE90}.

Before using the dataset, we performed sliding window cropping on each image. The cropped images were resized to a resolution of 1024x1024, with an overlap of 512 pixels between adjacent sub-images. To ensure experimental efficiency, we did not perform multi-scale scaling during the cropping process. All training and testing of the algorithms were conducted on a single V100 GPU, with the batch size of 2. The training process lasted for 12 epochs, and a learning rate warming-up strategy is employed to gradually increase the learning rate at the beginning of training.

\subsection{Ablation study}
\label{subsec:Ablation study}

In this section, we employ Oriented RCNN as the baseline algorithm and replaced the FPN and second-stage regression loss with our proposed Edge-FPN and Edge-Loss. We conducted experiments on the DOTA-v1.0 dataset and reported our results.

\subsubsection{Ablation study on Edge-FPN}

According to the edge self-attention mechanism shown in Figure 2, we designed two ways to utilize this mechanism:

Scheme 1: Only apply the edge self-attention to the input of the subsequent region proposal network.

Scheme 2: In order to retain the features before the self-attention mechanism, we perform feature fusion in the edge-FPN and the original features through a 1x1 convolution, and then input them to the region proposal network.


The experimental results of each scheme and the baseline on the DOTA dataset are shown in Table .\ref{tab:two schemel}. It can be observed that when using only the edge self-attention mechanism in the FPN, there is only a marginal improvement at the mAP evaluation metric. We speculate that is because the edge information not only contains the edges of the true objects but also may include a significant number of background edges in some cases, which leads the network to focus the attention on wrong areas. However, when performing feature fusion between the edge-FPN and the features before applying the attention, there is a significant improvement at mAP, indicating that no information loss occurred in the fusion process. Additionally, it can be observed that the improvement in average precision (AP) is mainly noticeable for regular-shaped objects. 

\begin{table}[htbp]
  \centering
  \caption{The experimental results of the two different edge self-attention mechanism usage schemes on the DOTA dataset}
    \begin{tabular}{lcccccc}
    \toprule
    \multicolumn{1}{c}{\multirow{2}[4]{*}{class}} & \multicolumn{2}{c}{FPN} & \multicolumn{2}{c}{Scheme 1} & \multicolumn{2}{c}{Scheme 2} \\
\cmidrule{2-7}          & recall & ap    & recall & ap    & recall & ap \\
    \midrule
    PL    & 0.93  & 0.895 & 0.934 & 0.896 & 0.932 & \textbf{0.897} \\
    BD    & 0.883 & \textbf{0.77} & 0.869 & 0.746 & 0.888 & 0.761 \\
    BR    & 0.709 & \textbf{0.544} & 0.704 & 0.516 & 0.699 & 0.538 \\
    GTF   & 0.906 & 0.764 & 0.925 & 0.771 & 0.911 & \textbf{0.775} \\
    SV    & 0.843 & 0.687 & 0.857 & \textbf{0.693} & 0.838 & 0.683 \\
    LV    & 0.917 & \textbf{0.842} & 0.909 & 0.838 & 0.913 & 0.841 \\
    SH    & 0.938 & 0.893 & 0.947 & 0.894 & 0.945 & \textbf{0.894} \\
    TC    & 0.936 & 0.908 & 0.944 & 0.907 & 0.938 & \textbf{0.908} \\
    BC    & 0.868 & 0.737 & 0.879 & 0.746 & 0.916 & \textbf{0.751} \\
    ST    & 0.686 & 0.626 & 0.701 & 0.655 & 0.704 & \textbf{0.692} \\
    SBF   & 0.837 & 0.632 & 0.869 & \textbf{0.633} & 0.829 & 0.616 \\
    RA    & 0.738 & 0.653 & 0.738 & 0.684 & 0.749 & \textbf{0.686} \\
    HA    & 0.823 & \textbf{0.747} & 0.814 & 0.743 & 0.819 & 0.736 \\
    SP    & 0.762 & 0.586 & 0.78  & 0.6   & 0.744 & \textbf{0.598} \\
    HC    & 0.705 & 0.588 & 0.697 & 0.572 & 0.688 & \textbf{0.643} \\
    \midrule
    mAP   &       & 0.725 &       & 0.726 &       & \textbf{0.735} \\
    \bottomrule
    \end{tabular}%
  \label{tab:two schemel}%
\end{table}%


We also compared the proposed edge self-attention mechanism with other attention mechanisms, such as Nonloacl2D \cite{wang2018non} and ContextBlock \cite{cao2019gcnet}. The results, as shown in Table \ref{tab:different seflf attention}, demonstrate the superiority of our proposed self-attention mechanism.

\begin{table}[htbp]
  \centering
  \caption{The performance of different self-attention mechanisms on the DOTA dataset}
    \begin{tabular}{cc}
    \toprule
    Method & mAP \\
    \midrule
    Nonloacl2D & 0.729 \\
    ContextBlock  & 0.731 \\
    Edge self-attention & 0.735 \\
    \bottomrule
    \end{tabular}%
  \label{tab:different seflf attention}%
\end{table}%

\subsubsection{Ablation study on Edge-Loss}
\label{Ablation study on Edge-Loss}

We conducted ablation experiments on the DOTA dataset to observe the individual impacts of the two innovative aspects of our Edge-Loss, via replacing the Smooth L1 Loss used in the second-stage regression branch of the baseline (Oriented RCNN). Our Edge-Loss introduces two main innovations:

1) Addressing the non-differentiability issue of the similarity function with respect to the five regression parameters of the PB by approximating the positions of edge points.

2) Improving the semantic mismatch issue between the GT box and PB through the use of relative positions and Gaussian weighting.

we compared the results of the ablation experiments, as shown in Table \ref{tab:innovative aspects}.

\begin{table}[]
  \centering
  \caption{The performances of different innovative aspects of Edge-Loss on the DOTA dataset. "Smooth L1" refers to the baseline's loss function, and "Origin" refers to the direct similarity function used in the template matching task, \textcircled{1} and \textcircled{2} mean the two innovative aspects illstrated in \ref{Ablation study on Edge-Loss}.}
    \begin{tabular}{c|cc|c}
    \toprule
          & \textcircled{1}     & \textcircled{2}     & mAP \\
    \midrule
    Smooth L1 &       &       & 0.725 \\
    \midrule
    Original &       &       & 0.526 \\
          & \checkmark     &       & 0.706 \\
          &       & \checkmark      & 0.727 \\
          & \checkmark     & \checkmark     & 0.731 \\
    \bottomrule
    \end{tabular}%
  \label{tab:innovative aspects}%
\end{table}%

\begin{figure*}[ht]
    \centering
    \includegraphics[width = 0.8\linewidth]{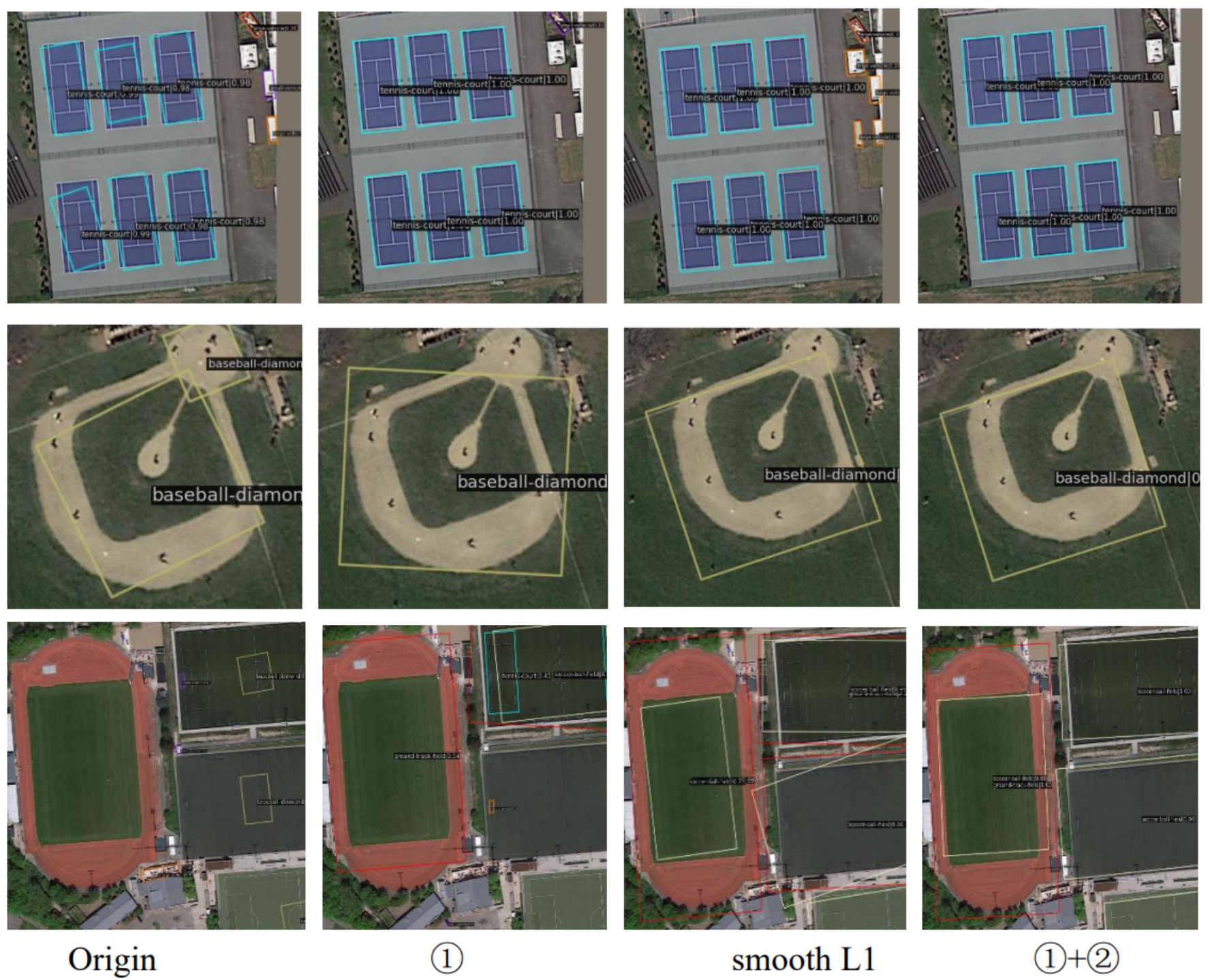}
    \caption{The visualized results about Edge-Loss on DOTA dataset}
    \label{fig:visualized results}
\end{figure*}

The results clearly indicate that when using the direct similarity function, the network's performance is poor. It is reason that the function is non-differentiable, making it challenging to optimize the five parameters of the PB. In this case, the regression of PB parameters heavily relies on the first-stage of the network, namely the RPN. After addressing the non-differentiability issue, the network's performance significantly improved, with an mAP of 0.706. However, it still falls short compared to the conventional Smooth L1 Loss. This is primarily due to the mismatch of gradient vectors caused by the different shapes and locations of the GT and PB. To address this issue, we introduced the use of relative positions and Gaussian weighting, which further improved the performance of our method. As a result, we achieved an mAP score of 0.731, surpassing the baseline's mAP of 0.725. We used the trained weights obtained from the aforementioned ablation experiments to perform predictions on the DOTA validation set. The visualized results are shown in Fig \ref{fig:visualized results}.

\subsection{Benchmark results}
\label{benchmark}

We conducted a comprehensive comparison with other one-stage and two-stage algorithms on the DOTA dataset. The results are shown in Table \ref{tab:all result}. It can be observed that our proposed algorithm exhibits significant improvements on AP for most regular objects, such as $PL, BD, GTF, and TC$, compared to other methods.

\begin{table*}[ht]
  \centering
  \caption{Detection results on DOTA dataset. Oriented RCNN + L  and Oriented RCNN + A means Oriented RCNN + Edge-Loss and Oriented RCNN + Edge self-attention respectively}
    \resizebox{\linewidth}{!}{
    
    \begin{tabular}{lcccccccccccccccc}
    \toprule
    Method & PL    & BD    & BR    & GTF   & SV    & LV    & SH    & TC    & BC    & ST    & SBF   & RA    & HA    & SP    & HC    & mAP \\
    \midrule
    RRPN  & 0.885  & 0.712  & 0.317  & 0.593  & 0.519  & 0.562  & 0.573  & 0.908  & 0.728  & \textbf{0.674 } & 0.567  & 0.528  & 0.531  & 0.519  & 0.536  & 0.610  \\
    R3Det & 0.905  & 0.701  & 0.473  & 0.691  & 0.700  & 0.801  & 0.867  & 0.909  & 0.729  & 0.580  & 0.537  & 0.646  & 0.625  & 0.548  & 0.473  & 0.679  \\
    S2ANet & 0.907  & 0.723  & 0.572  & 0.713  & 0.716  & 0.860  & 0.889  & 0.909  & 0.749  & 0.655  & 0.620  & 0.710  & 0.653  & 0.578  & 0.627  & 0.725  \\
    \midrule
    \midrule
    Rotated Faster R-CNN & 0.904  & 0.732  & 0.524  & 0.733  & 0.739  & 0.855  & 0.888  & 0.909  & 0.752  & 0.594  & 0.689  & 0.657  & 0.657  & 0.615  & 0.647  & 0.726  \\
    ReDet & 0.907  & 0.746  & 0.598  & 0.730  & 0.726  & 0.881  & 0.899  & 0.909  & 0.751  & 0.600  & 0.715  & 0.654  & 0.730  & 0.596  & 0.704  & 0.743  \\
    RoI Trans & 0.907  & 0.742  & \textbf{0.616 } & 0.740  & \textbf{0.743 } & \textbf{0.884 } & \textbf{0.899 } & 0.909  & \textbf{0.813 } & 0.600  & \textbf{0.719 } & 0.714  & 0.736  & 0.630  & \textbf{0.725 } & 0.759  \\
    \midrule
    \midrule
    Oriented RCNN  & 0.895  & 0.810  & 0.573  & 0.804  & 0.687  & 0.842  & 0.893  & 0.908  & 0.776  & 0.659  & 0.665  & 0.687  & 0.786  & 0.617  & 0.619  & 0.748  \\
    Oriented RCNN+A & 0.897  & 0.801  & 0.566  & 0.816  & 0.683  & 0.841  & 0.894  & 0.908  & 0.790  & 0.671  & 0.648  & \textbf{0.722 } & 0.775  & 0.629  & 0.677  & 0.755  \\
    Oriented RCNN +L & 0.906  & 0.829  & 0.572  & 0.814  & 0.681  & 0.861  & 0.887  & 0.918  & 0.799  & 0.670  & 0.680  & 0.681  & 0.776  & 0.630  & 0.612  & 0.754  \\
    Oriented RCNN +A+L & \textbf{0.907 } & \textbf{0.830 } & 0.584  & \textbf{0.824 } & 0.689  & 0.868  & 0.898  & \textbf{0.920 } & 0.804  & 0.673  & 0.692  & 0.686  & \textbf{0.788 } & \textbf{0.633 } & 0.621  & \textbf{0.761 } \\
    \bottomrule
    \end{tabular}%
  \label{tab:addlabel}%
    }
  \label{tab:all result}%
\end{table*}%

\section{Conclusion}
\label{sec:conc}
Firstly, we propose a unique loss function called Edge-Loss, which is based on the similarity measurement of edge gradient vectors, inspired by the shape-based template matching task. We address the issues of non-differentiability and semantic misalignment between the gradient vectors in GT and PB to make the loss function workable and more superior. Secondly, we introduce an innovative attention mechanism based on the object edges, aiming to guide the detection network to focus more on the object edges. We validate the effectiveness of these two innovations on the DOTA dataset. Compared to the baseline method, Oriented RCNN, our approach achieves a significant improvement of 1.3 percentage points in terms of the mAP evaluation metric.

\bibliographystyle{cas-model2-names}

\bibliography{cas-refs}

\begin{thebibliography}{34}
\expandafter\ifx\csname natexlab\endcsname\relax\def\natexlab#1{#1}\fi
\providecommand{\url}[1]{\texttt{#1}}
\providecommand{\href}[2]{#2}
\providecommand{\path}[1]{#1}
\providecommand{\DOIprefix}{doi:}
\providecommand{\ArXivprefix}{arXiv:}
\providecommand{\URLprefix}{URL: }
\providecommand{\Pubmedprefix}{pmid:}
\providecommand{\doi}[1]{\href{http://dx.doi.org/#1}{\path{#1}}}
\providecommand{\Pubmed}[1]{\href{pmid:#1}{\path{#1}}}
\providecommand{\bibinfo}[2]{#2}
\ifx\xfnm\relax \def\xfnm[#1]{\unskip,\space#1}\fi
\bibitem[{Cao et~al.(2019)Cao, Xu, Lin, Wei and Hu}]{cao2019gcnet}
\bibinfo{author}{Cao, Y.}, \bibinfo{author}{Xu, J.}, \bibinfo{author}{Lin, S.}, \bibinfo{author}{Wei, F.}, \bibinfo{author}{Hu, H.}, \bibinfo{year}{2019}.
\newblock \bibinfo{title}{Gcnet: Non-local networks meet squeeze-excitation networks and beyond}, in: \bibinfo{booktitle}{Proceedings of the IEEE/CVF international conference on computer vision workshops}, pp. \bibinfo{pages}{0--0}.
\bibitem[{Chen et~al.(2020)Chen, Chen, Lin, See, Yu, Ke and Yang}]{chen2020piou}
\bibinfo{author}{Chen, Z.}, \bibinfo{author}{Chen, K.}, \bibinfo{author}{Lin, W.}, \bibinfo{author}{See, J.}, \bibinfo{author}{Yu, H.}, \bibinfo{author}{Ke, Y.}, \bibinfo{author}{Yang, C.}, \bibinfo{year}{2020}.
\newblock \bibinfo{title}{Piou loss: Towards accurate oriented object detection in complex environments}, in: \bibinfo{booktitle}{Computer Vision--ECCV 2020: 16th European Conference, Glasgow, UK, August 23--28, 2020, Proceedings, Part V 16}, \bibinfo{organization}{Springer}. pp. \bibinfo{pages}{195--211}.
\bibitem[{Crispin and Rankov(2007)}]{ncc}
\bibinfo{author}{Crispin, A.}, \bibinfo{author}{Rankov, V.}, \bibinfo{year}{2007}.
\newblock \bibinfo{title}{Automated inspection of pcb components using a genetic algorithm template-matching approach}.
\newblock \bibinfo{journal}{The International Journal of Advanced Manufacturing Technology} \bibinfo{volume}{35}, \bibinfo{pages}{293--300}.
\bibitem[{Dekel et~al.(2015)Dekel, Oron, Rubinstein, Avidan and Freeman}]{dekel2015best}
\bibinfo{author}{Dekel, T.}, \bibinfo{author}{Oron, S.}, \bibinfo{author}{Rubinstein, M.}, \bibinfo{author}{Avidan, S.}, \bibinfo{author}{Freeman, W.T.}, \bibinfo{year}{2015}.
\newblock \bibinfo{title}{Best-buddies similarity for robust template matching}, in: \bibinfo{booktitle}{Proceedings of the IEEE conference on computer vision and pattern recognition}, pp. \bibinfo{pages}{2021--2029}.
\bibitem[{Ding et~al.(2019)Ding, Xue, Long, Xia and Lu}]{ding2019learning}
\bibinfo{author}{Ding, J.}, \bibinfo{author}{Xue, N.}, \bibinfo{author}{Long, Y.}, \bibinfo{author}{Xia, G.S.}, \bibinfo{author}{Lu, Q.}, \bibinfo{year}{2019}.
\newblock \bibinfo{title}{Learning roi transformer for oriented object detection in aerial images}, in: \bibinfo{booktitle}{Proceedings of the IEEE/CVF Conference on Computer Vision and Pattern Recognition}, pp. \bibinfo{pages}{2849--2858}.
\bibitem[{Han et~al.(2021a)Han, Ding, Li and Xia}]{han2021align}
\bibinfo{author}{Han, J.}, \bibinfo{author}{Ding, J.}, \bibinfo{author}{Li, J.}, \bibinfo{author}{Xia, G.S.}, \bibinfo{year}{2021}a.
\newblock \bibinfo{title}{Align deep features for oriented object detection}.
\newblock \bibinfo{journal}{IEEE Transactions on Geoscience and Remote Sensing} \bibinfo{volume}{60}, \bibinfo{pages}{1--11}.
\bibitem[{Han et~al.(2021b)Han, Ding, Xue and Xia}]{han2021redet}
\bibinfo{author}{Han, J.}, \bibinfo{author}{Ding, J.}, \bibinfo{author}{Xue, N.}, \bibinfo{author}{Xia, G.S.}, \bibinfo{year}{2021}b.
\newblock \bibinfo{title}{Redet: A rotation-equivariant detector for aerial object detection}, in: \bibinfo{booktitle}{Proceedings of the IEEE/CVF Conference on Computer Vision and Pattern Recognition}, pp. \bibinfo{pages}{2786--2795}.
\bibitem[{Hu et~al.(2018a)Hu, Shen, Albanie, Sun and Vedaldi}]{hu2018gather}
\bibinfo{author}{Hu, J.}, \bibinfo{author}{Shen, L.}, \bibinfo{author}{Albanie, S.}, \bibinfo{author}{Sun, G.}, \bibinfo{author}{Vedaldi, A.}, \bibinfo{year}{2018}a.
\newblock \bibinfo{title}{Gather-excite: Exploiting feature context in convolutional neural networks}.
\newblock \bibinfo{journal}{Advances in neural information processing systems} \bibinfo{volume}{31}.
\bibitem[{Hu et~al.(2018b)Hu, Shen and Sun}]{hu2018squeeze}
\bibinfo{author}{Hu, J.}, \bibinfo{author}{Shen, L.}, \bibinfo{author}{Sun, G.}, \bibinfo{year}{2018}b.
\newblock \bibinfo{title}{Squeeze-and-excitation networks}, in: \bibinfo{booktitle}{Proceedings of the IEEE conference on computer vision and pattern recognition}, pp. \bibinfo{pages}{7132--7141}.
\bibitem[{Jaderberg et~al.(2015)Jaderberg, Simonyan, Zisserman et~al.}]{jaderberg2015spatial}
\bibinfo{author}{Jaderberg, M.}, \bibinfo{author}{Simonyan, K.}, \bibinfo{author}{Zisserman, A.}, et~al., \bibinfo{year}{2015}.
\newblock \bibinfo{title}{Spatial transformer networks}.
\newblock \bibinfo{journal}{Advances in neural information processing systems} \bibinfo{volume}{28}.
\bibitem[{Lan et~al.(2021)Lan, Wu and Li}]{lan2021gad}
\bibinfo{author}{Lan, Y.}, \bibinfo{author}{Wu, X.}, \bibinfo{author}{Li, Y.}, \bibinfo{year}{2021}.
\newblock \bibinfo{title}{Gad: A global-aware diversity-based template matching method}.
\newblock \bibinfo{journal}{IEEE Transactions on Instrumentation and Measurement} \bibinfo{volume}{71}, \bibinfo{pages}{1--13}.
\bibitem[{Li et~al.(2022)Li, Chen, Hu and Zhu}]{li2022oriented}
\bibinfo{author}{Li, W.}, \bibinfo{author}{Chen, Y.}, \bibinfo{author}{Hu, K.}, \bibinfo{author}{Zhu, J.}, \bibinfo{year}{2022}.
\newblock \bibinfo{title}{Oriented reppoints for aerial object detection}, in: \bibinfo{booktitle}{Proceedings of the IEEE/CVF conference on computer vision and pattern recognition}, pp. \bibinfo{pages}{1829--1838}.
\bibitem[{Li et~al.(2019)Li, Wang, Hu and Yang}]{li2019selective}
\bibinfo{author}{Li, X.}, \bibinfo{author}{Wang, W.}, \bibinfo{author}{Hu, X.}, \bibinfo{author}{Yang, J.}, \bibinfo{year}{2019}.
\newblock \bibinfo{title}{Selective kernel networks}, in: \bibinfo{booktitle}{Proceedings of the IEEE/CVF conference on computer vision and pattern recognition}, pp. \bibinfo{pages}{510--519}.
\bibitem[{Li et~al.(2023)Li, Hou, Zheng, Cheng, Yang and Li}]{li2023large}
\bibinfo{author}{Li, Y.}, \bibinfo{author}{Hou, Q.}, \bibinfo{author}{Zheng, Z.}, \bibinfo{author}{Cheng, M.M.}, \bibinfo{author}{Yang, J.}, \bibinfo{author}{Li, X.}, \bibinfo{year}{2023}.
\newblock \bibinfo{title}{Large selective kernel network for remote sensing object detection}.
\newblock \bibinfo{journal}{arXiv preprint arXiv:2303.09030} .
\bibitem[{Lin et~al.(2017)Lin, Goyal, Girshick, He and Doll{\'a}r}]{lin2017focal}
\bibinfo{author}{Lin, T.Y.}, \bibinfo{author}{Goyal, P.}, \bibinfo{author}{Girshick, R.}, \bibinfo{author}{He, K.}, \bibinfo{author}{Doll{\'a}r, P.}, \bibinfo{year}{2017}.
\newblock \bibinfo{title}{Focal loss for dense object detection}, in: \bibinfo{booktitle}{Proceedings of the IEEE international conference on computer vision}, pp. \bibinfo{pages}{2980--2988}.
\bibitem[{Lin et~al.(2014)Lin, Maire, Belongie, Hays, Perona, Ramanan, Doll{\'a}r and Zitnick}]{lin2014microsoft}
\bibinfo{author}{Lin, T.Y.}, \bibinfo{author}{Maire, M.}, \bibinfo{author}{Belongie, S.}, \bibinfo{author}{Hays, J.}, \bibinfo{author}{Perona, P.}, \bibinfo{author}{Ramanan, D.}, \bibinfo{author}{Doll{\'a}r, P.}, \bibinfo{author}{Zitnick, C.L.}, \bibinfo{year}{2014}.
\newblock \bibinfo{title}{Microsoft coco: Common objects in context}, in: \bibinfo{booktitle}{Computer Vision--ECCV 2014: 13th European Conference, Zurich, Switzerland, September 6-12, 2014, Proceedings, Part V 13}, \bibinfo{organization}{Springer}. pp. \bibinfo{pages}{740--755}.
\bibitem[{Lin et~al.(2021)Lin, Guo and Wang}]{lin2021global}
\bibinfo{author}{Lin, X.}, \bibinfo{author}{Guo, Y.a.}, \bibinfo{author}{Wang, J.}, \bibinfo{year}{2021}.
\newblock \bibinfo{title}{Global correlation network: End-to-end joint multi-object detection and tracking}.
\newblock \bibinfo{journal}{arXiv preprint arXiv:2103.12511} .
\bibitem[{Liu et~al.(2016)Liu, Wang, Weng and Yang}]{liu2016ship}
\bibinfo{author}{Liu, Z.}, \bibinfo{author}{Wang, H.}, \bibinfo{author}{Weng, L.}, \bibinfo{author}{Yang, Y.}, \bibinfo{year}{2016}.
\newblock \bibinfo{title}{Ship rotated bounding box space for ship extraction from high-resolution optical satellite images with complex backgrounds}.
\newblock \bibinfo{journal}{IEEE geoscience and remote sensing letters} \bibinfo{volume}{13}, \bibinfo{pages}{1074--1078}.
\bibitem[{Lu et~al.(2021)Lu, Zhang, Pang, Li and Zhu}]{lu2021robust}
\bibinfo{author}{Lu, Y.}, \bibinfo{author}{Zhang, X.}, \bibinfo{author}{Pang, S.}, \bibinfo{author}{Li, H.}, \bibinfo{author}{Zhu, B.}, \bibinfo{year}{2021}.
\newblock \bibinfo{title}{A robust edge-based template matching algorithm for displacement measurement of compliant mechanisms under scanning electron microscope}.
\newblock \bibinfo{journal}{Review of Scientific Instruments} \bibinfo{volume}{92}.
\bibitem[{Pan et~al.(2020)Pan, Ren, Sheng, Dong, Yuan, Guo, Ma and Xu}]{pan2020dynamic}
\bibinfo{author}{Pan, X.}, \bibinfo{author}{Ren, Y.}, \bibinfo{author}{Sheng, K.}, \bibinfo{author}{Dong, W.}, \bibinfo{author}{Yuan, H.}, \bibinfo{author}{Guo, X.}, \bibinfo{author}{Ma, C.}, \bibinfo{author}{Xu, C.}, \bibinfo{year}{2020}.
\newblock \bibinfo{title}{Dynamic refinement network for oriented and densely packed object detection}, in: \bibinfo{booktitle}{Proceedings of the IEEE/CVF Conference on Computer Vision and Pattern Recognition}, pp. \bibinfo{pages}{11207--11216}.
\bibitem[{Steger(2002)}]{steger2002occlusion}
\bibinfo{author}{Steger, C.}, \bibinfo{year}{2002}.
\newblock \bibinfo{title}{Occlusion, clutter, and illumination invariant object recognition}.
\newblock \bibinfo{journal}{International Archives of Photogrammetry Remote Sensing and Spatial Information Sciences} \bibinfo{volume}{34}, \bibinfo{pages}{345--350}.
\bibitem[{Swaroop and Sharma(2016)}]{overview2016}
\bibinfo{author}{Swaroop, P.}, \bibinfo{author}{Sharma, N.}, \bibinfo{year}{2016}.
\newblock \bibinfo{title}{An overview of various template matching methodologies in image processing}.
\newblock \bibinfo{journal}{International Journal of Computer Applications} \bibinfo{volume}{153}, \bibinfo{pages}{8--14}.
\bibitem[{Talker et~al.(2018)Talker, Moses and Shimshoni}]{talker2018efficient}
\bibinfo{author}{Talker, L.}, \bibinfo{author}{Moses, Y.}, \bibinfo{author}{Shimshoni, I.}, \bibinfo{year}{2018}.
\newblock \bibinfo{title}{Efficient sliding window computation for nn-based template matching}, in: \bibinfo{booktitle}{Proceedings of the European Conference on Computer Vision (ECCV)}, pp. \bibinfo{pages}{404--418}.
\bibitem[{Talmi et~al.(2017)Talmi, Mechrez and Zelnik-Manor}]{talmi2017template}
\bibinfo{author}{Talmi, I.}, \bibinfo{author}{Mechrez, R.}, \bibinfo{author}{Zelnik-Manor, L.}, \bibinfo{year}{2017}.
\newblock \bibinfo{title}{Template matching with deformable diversity similarity}, in: \bibinfo{booktitle}{Proceedings of the IEEE Conference on Computer Vision and Pattern Recognition}, pp. \bibinfo{pages}{175--183}.
\bibitem[{Wang et~al.(2020)Wang, Wu, Zhu, Li, Zuo and Hu}]{wang2020eca}
\bibinfo{author}{Wang, Q.}, \bibinfo{author}{Wu, B.}, \bibinfo{author}{Zhu, P.}, \bibinfo{author}{Li, P.}, \bibinfo{author}{Zuo, W.}, \bibinfo{author}{Hu, Q.}, \bibinfo{year}{2020}.
\newblock \bibinfo{title}{Eca-net: Efficient channel attention for deep convolutional neural networks}, in: \bibinfo{booktitle}{Proceedings of the IEEE/CVF conference on computer vision and pattern recognition}, pp. \bibinfo{pages}{11534--11542}.
\bibitem[{Wang et~al.(2018)Wang, Girshick, Gupta and He}]{wang2018non}
\bibinfo{author}{Wang, X.}, \bibinfo{author}{Girshick, R.}, \bibinfo{author}{Gupta, A.}, \bibinfo{author}{He, K.}, \bibinfo{year}{2018}.
\newblock \bibinfo{title}{Non-local neural networks}, in: \bibinfo{booktitle}{Proceedings of the IEEE conference on computer vision and pattern recognition}, pp. \bibinfo{pages}{7794--7803}.
\bibitem[{Xia et~al.(2018)Xia, Bai, Ding, Zhu, Belongie, Luo, Datcu, Pelillo and Zhang}]{xia2018dota}
\bibinfo{author}{Xia, G.S.}, \bibinfo{author}{Bai, X.}, \bibinfo{author}{Ding, J.}, \bibinfo{author}{Zhu, Z.}, \bibinfo{author}{Belongie, S.}, \bibinfo{author}{Luo, J.}, \bibinfo{author}{Datcu, M.}, \bibinfo{author}{Pelillo, M.}, \bibinfo{author}{Zhang, L.}, \bibinfo{year}{2018}.
\newblock \bibinfo{title}{Dota: A large-scale dataset for object detection in aerial images}, in: \bibinfo{booktitle}{Proceedings of the IEEE conference on computer vision and pattern recognition}, pp. \bibinfo{pages}{3974--3983}.
\bibitem[{Xie et~al.(2021)Xie, Cheng, Wang, Yao and Han}]{xie2021oriented}
\bibinfo{author}{Xie, X.}, \bibinfo{author}{Cheng, G.}, \bibinfo{author}{Wang, J.}, \bibinfo{author}{Yao, X.}, \bibinfo{author}{Han, J.}, \bibinfo{year}{2021}.
\newblock \bibinfo{title}{Oriented r-cnn for object detection}, in: \bibinfo{booktitle}{Proceedings of the IEEE/CVF international conference on computer vision}, pp. \bibinfo{pages}{3520--3529}.
\bibitem[{Yang et~al.(2021a)Yang, Yan, Feng and He}]{yang2021r3det}
\bibinfo{author}{Yang, X.}, \bibinfo{author}{Yan, J.}, \bibinfo{author}{Feng, Z.}, \bibinfo{author}{He, T.}, \bibinfo{year}{2021}a.
\newblock \bibinfo{title}{R3det: Refined single-stage detector with feature refinement for rotating object}, in: \bibinfo{booktitle}{Proceedings of the AAAI conference on artificial intelligence}, pp. \bibinfo{pages}{3163--3171}.
\bibitem[{Yang et~al.(2021b)Yang, Yan, Ming, Wang, Zhang and Tian}]{yang2021rethinking}
\bibinfo{author}{Yang, X.}, \bibinfo{author}{Yan, J.}, \bibinfo{author}{Ming, Q.}, \bibinfo{author}{Wang, W.}, \bibinfo{author}{Zhang, X.}, \bibinfo{author}{Tian, Q.}, \bibinfo{year}{2021}b.
\newblock \bibinfo{title}{Rethinking rotated object detection with gaussian wasserstein distance loss}, in: \bibinfo{booktitle}{International conference on machine learning}, \bibinfo{organization}{PMLR}. pp. \bibinfo{pages}{11830--11841}.
\bibitem[{Yang et~al.(2019)Yang, Yang, Yan, Zhang, Zhang, Guo, Sun and Fu}]{yang2019scrdet}
\bibinfo{author}{Yang, X.}, \bibinfo{author}{Yang, J.}, \bibinfo{author}{Yan, J.}, \bibinfo{author}{Zhang, Y.}, \bibinfo{author}{Zhang, T.}, \bibinfo{author}{Guo, Z.}, \bibinfo{author}{Sun, X.}, \bibinfo{author}{Fu, K.}, \bibinfo{year}{2019}.
\newblock \bibinfo{title}{Scrdet: Towards more robust detection for small, cluttered and rotated objects}, in: \bibinfo{booktitle}{Proceedings of the IEEE/CVF international conference on computer vision}, pp. \bibinfo{pages}{8232--8241}.
\bibitem[{Yang et~al.(2021c)Yang, Yang, Yang, Ming, Wang, Tian and Yan}]{yang2021learning}
\bibinfo{author}{Yang, X.}, \bibinfo{author}{Yang, X.}, \bibinfo{author}{Yang, J.}, \bibinfo{author}{Ming, Q.}, \bibinfo{author}{Wang, W.}, \bibinfo{author}{Tian, Q.}, \bibinfo{author}{Yan, J.}, \bibinfo{year}{2021}c.
\newblock \bibinfo{title}{Learning high-precision bounding box for rotated object detection via kullback-leibler divergence}.
\newblock \bibinfo{journal}{Advances in Neural Information Processing Systems} \bibinfo{volume}{34}, \bibinfo{pages}{18381--18394}.
\bibitem[{Yin et~al.(2018)Yin, Pan, Su, Liu and Peng}]{yin2018fast}
\bibinfo{author}{Yin, J.}, \bibinfo{author}{Pan, H.}, \bibinfo{author}{Su, H.}, \bibinfo{author}{Liu, Z.}, \bibinfo{author}{Peng, Z.}, \bibinfo{year}{2018}.
\newblock \bibinfo{title}{A fast orientation invariant detector based on the one-stage method}, in: \bibinfo{booktitle}{MATEC Web of Conferences}, \bibinfo{organization}{EDP Sciences}. p. \bibinfo{pages}{04036}.
\bibitem[{Zheng et~al.(2020)Zheng, Zhang, Xie, Lu and Zhou}]{zheng2020rotation}
\bibinfo{author}{Zheng, Y.}, \bibinfo{author}{Zhang, D.}, \bibinfo{author}{Xie, S.}, \bibinfo{author}{Lu, J.}, \bibinfo{author}{Zhou, J.}, \bibinfo{year}{2020}.
\newblock \bibinfo{title}{Rotation-robust intersection over union for 3d object detection}, in: \bibinfo{booktitle}{European Conference on Computer Vision}, \bibinfo{organization}{Springer}. pp. \bibinfo{pages}{464--480}.

\end{thebibliography}

\bio{}

\endbio

\end{document}